\documentclass[10pt,twocolumn,letterpaper]{article}

\usepackage[pagenumbers]{cvpr}  

\usepackage{CJKutf8}

\usepackage{placeins}

\usepackage{graphicx}
\usepackage{booktabs}
\usepackage{colortbl}
\usepackage{amssymb}
\usepackage{subcaption}
\usepackage{algorithm}
\usepackage{algpseudocode}
\usepackage{float}
\usepackage{CJKutf8}
\usepackage{xspace}

\definecolor{OursRow}{RGB}{244,241,252}
\newcommand{\ourmethod}{PhaseOPD}

\usepackage{amsmath,amsfonts,bm}

\def\eqref#1{equation~\ref{#1}}

\def\1{\bm{1}}

\DeclareMathAlphabet{\mathsfit}{\encodingdefault}{\sfdefault}{m}{sl}
\SetMathAlphabet{\mathsfit}{bold}{\encodingdefault}{\sfdefault}{bx}{n}

\renewcommand{\eqref}[1]{\textup{(\ref{#1})}}

\definecolor{cvprblue}{rgb}{0.21,0.49,0.74}
\usepackage[pagebackref,breaklinks,colorlinks,allcolors=cvprblue]{hyperref}

\def\paperID{*****}
\def\confName{CVPR}
\def\confYear{2026}

\title{Timing Matters Twice: When to Query and When to Skip in Flow-Matching On-Policy Distillation.}

\author{%
Renye Yan$^{1,*}$\quad
Junyi Zhan$^{1,*}$\quad
Jikang Cheng$^{1,*,\dagger}$\quad
Wei Peng$^{2}$\\[4pt]
$^{1}$Peking University\qquad $^{2}$Stanford University\\[3pt]
{\small $^{*}$Equal contribution.\qquad $^{\dagger}$Project lead.}%
}
\hypersetup{pdfauthor={Renye Yan, Junyi Zhan, Jikang Cheng, Wei Peng}}

\begin{document}
\begin{CJK*}{UTF8}{gbsn}
\maketitle

\begin{abstract}
On-policy distillation (OPD) enables flow-matching models to acquire complementary capabilities from specialized teachers through supervision on student-generated trajectories. However, different evaluation dimensions (e.g., aesthetics and OCR accuracy) and tasks (e.g., image generation and editing) emphasize different aspects of the denoising process, suggesting that their effective supervision may occur at different stages. Existing methods typically select supervision states from predefined timestep distributions, which may overlook these phase-dependent differences with unnecessary computation overhead. This motivates two critical timing decisions: when to adaptively query teachers for effective supervision based on the evolving generative process, and when to skip the remaining rollout to avoid redundant computation, \textit{i.e.}, \textit{timing matters twice}. 
In this paper, we introduce \textit{\textbf{PhaseOPD}}, a phase-aware query-and-skip framework that identifies generative phases online from revisions of predicted-clean latents, using existing student velocity predictions without additional model evaluations. For each task-assigned teacher, PhaseOPD selects a supervision state based on its phase preference and skips the unused rollout suffix once all query decisions are finalized, preserving the single-query velocity-distillation objective.
On SD3.5-M, PhaseOPD executes only 5.56 of 16 rollout positions per local batch (65.2\% fewer), reducing training time by 41.3\% relative to DanceOPD while improving Aesthetic, GenEval, and OCR scores.

\end{abstract}

\section{Introduction}

\begin{figure}[!t]
    \centering
    \includegraphics[width=\columnwidth]{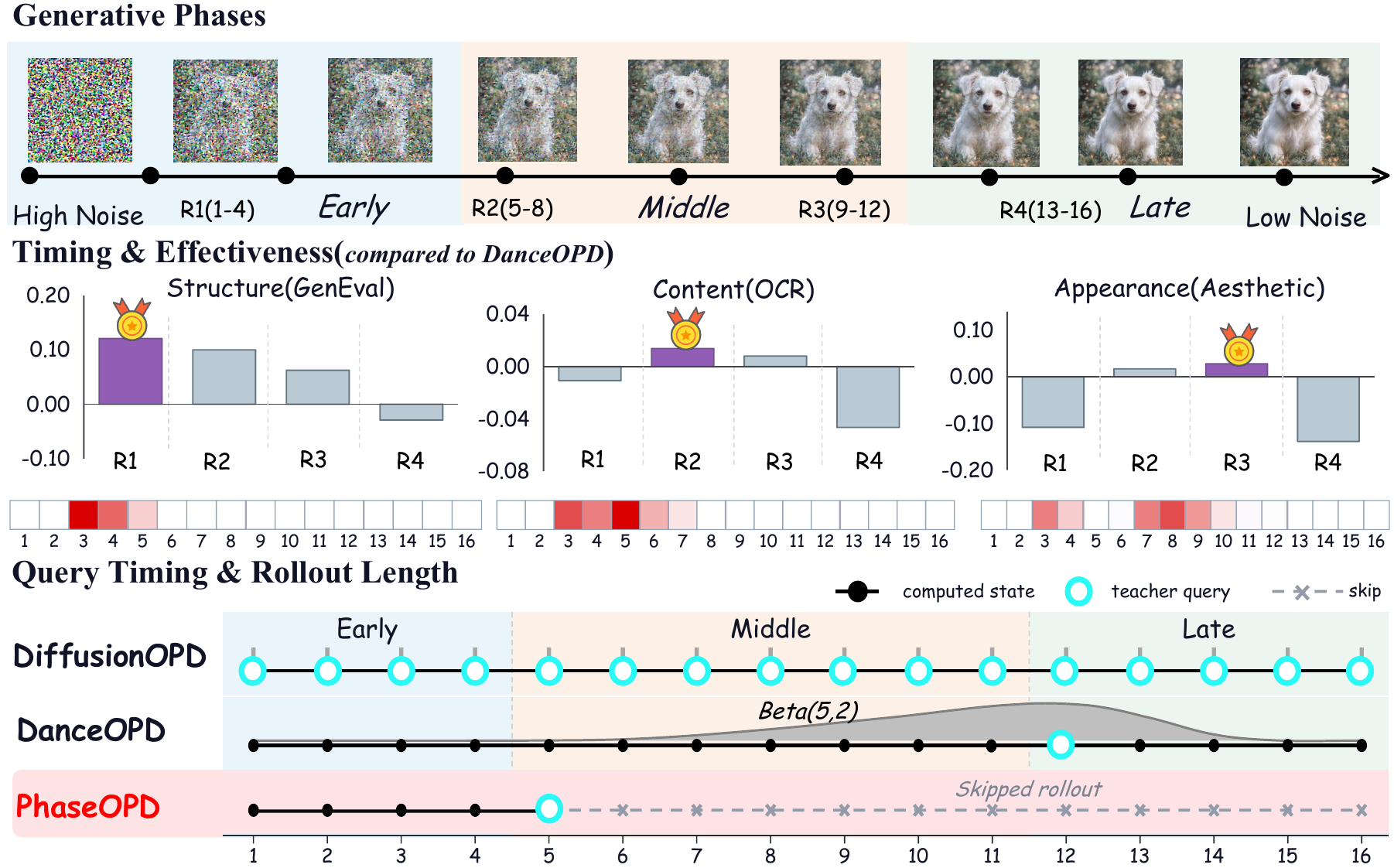}
    \caption{Teacher supervision is not equally useful throughout denoising: semantic composition, text rendering, and aesthetics benefit most from different query intervals. This motivates adapting query timing to each teacher-task pair and terminating the rollout after the required queries, rather than using a shared preset schedule and completing every denoising step.}
    \label{fig:motivation}
\end{figure}

In recent years, diffusion and flow-matching models have made significant progress in image generation~\cite{li2026videococo,ling2026gentrack,ma2026can,yan2026swiftexplorer,yan2026beyond}. The research focus has gradually shifted from improving basic generative capabilities to aligning models with challenging task-specific objectives and preferences through post-training. In complex applications, generative models must satisfy multiple criteria simultaneously, including accurate semantic composition, coherent spatial structure, high-quality local details, and visual appearance aligned with human preferences. Improving multiple visual objectives within a unified model has therefore become an important challenge in generative model post-training. Previous studies have mainly combined reward signals from different evaluation dimensions using weighted aggregation and optimized the generation distribution of pretrained models through reinforcement learning~\cite{zhao2022alphaholdem,yan2026past}. However, such methods not only rely on sparse scalar rewards for optimization, but also suffer from competing update directions induced by heterogeneous rewards, often resulting in gradient interference and imbalanced capabilities.

To mitigate reward sparsity and gradient conflicts in joint multi-objective optimization, recent studies have introduced On-Policy Distillation (OPD) into generative models \citep{fang2026flowopd,li2026diffusionopd,zhou2026danceopd}. Instead of directly optimizing a single policy with multiple rewards, multi-teacher OPD first trains specialized teachers independently for different objectives. These teachers then provide supervision on the student’s own generation trajectories, integrating diverse generative capabilities into a unified student model.

Although OPD provides an effective approach to multi-objective optimization, existing OPD paradigms do not adequately characterize how the value of each task-specific teacher's supervision changes across generation stages. Because each teacher is trained for a distinct objective, the key question is which part of the student's trajectory provides the most effective supervision for that teacher and its corresponding task—not which teacher should be assigned to a given stage. This raises two connected questions: for each teacher-task pair, at what stage should the student be queried to obtain the most useful supervision? Once the relevant query for each task has been completed, when can the remaining rollout be terminated to avoid redundant computation?
When query timing is not specified for each teacher-task pair, it remains unclear whether a chosen supervision stage is well suited to that teacher's objective. This matters because denoising follows a stage-dependent pattern: early stages focus on constructing global semantics, intermediate stages gradually shift toward local content generation, and later stages place greater emphasis on refining details and visual aesthetics. These evolving responsibilities mean that the utility of a given teacher's guidance depends on whether the student's state at a particular stage is suitable for that teacher's corresponding objective. Thus, supervision timing should be determined for each teacher-task pair, rather than by assigning teachers to stages in isolation.
Beyond effectively scheduling teacher supervision, avoiding unnecessary computational overhead is another important challenge for multi-teacher OPD. Obtaining teacher supervision requires long and continuous rollouts of student trajectories, which introduces considerable computational costs. More importantly, once all necessary teacher queries have been completed, or once the denoising process is close to convergence and the latent representation has become nearly stationary, continuing the remaining rollout provides little additional information for distillation. It may even introduce optimization conflicts due to a mismatch between teacher supervision and generation dynamics. Therefore, skipping redundant trajectory supervision while preserving its effectiveness is crucial to improving OPD training efficiency. Appendix.~\ref{sec:related_work} discusses related work in detail.

Building on these insights, we propose PhaseOPD, a dual timing-adaptive policy distillation framework that uses the student's generation dynamics to schedule each task-specific teacher at the stages where its supervision is most useful. Unlike existing methods that focus primarily on task-level teacher selection without jointly optimizing teacher-query timing and trajectory length, PhaseOPD identifies a suitable query interval for each teacher-task pair from the student's evolving trajectory.
For example, it schedules structure-oriented supervision during global layout and primary structure formation, supervision for objectives involving both structure and detail during the transition toward semantic content and salient details, and visual-quality supervision during later texture and appearance refinement. The task-specific teacher defines what is supervised, while the trajectory dynamics determine when its supervision is applied. This constitutes the first level of adaptivity in PhaseOPD.
In addition, PhaseOPD uses generation dynamics and the marginal utility of task-specific supervision to detect diminishing returns. It terminates the remaining rollout once no further teacher query is needed, thereby reducing computational overhead. This constitutes the second level of adaptivity in PhaseOPD and follows Occam’s razor: introduce supervision only when necessary.

In summary, our main contributions are threefold: (1) We revisit teacher organization in multi-teacher on-policy distillation by introducing a generation-dynamics dimension orthogonal to the task dimension. We further identify the mismatch between teacher supervision and denoising dynamics, a problem that task-level OPD cannot adequately characterize. (2) We propose PhaseOPD, which uses the student's dynamics to identify, for each task-specific teacher, the generation intervals where its supervision is most valuable and schedules queries accordingly. This reduces cross-stage interference and unnecessary over-constraint. (3) We introduce a dynamic truncation strategy that immediately terminates the subsequent rollout once the supervision signals determine the required teacher-query positions, avoiding redundant computation.

\section{Motivation}
\label{motivation}
Mainstream OPD paradigms overlook that different specialized teacher capabilities should align with the characteristics of different denoising stages. This oversight can create temporal conflicts between teacher supervision and denoising dynamics. If a teacher specializing in local details intervenes before the global structure stabilizes, its supervision may prematurely steer the model toward local optimization objectives and disrupt primary structure formation. Meanwhile, \textbf{because local details rely on an established global structure as their underlying carrier, the absence of a stable structural foundation also prevents detail-oriented supervision from fully realizing its value}. Conversely, once the primary structure has converged and the model has entered the local refinement stage, continued strong structural supervision may unnecessarily constrain subsequent detail enhancement. More generally, even if a teacher’s objective is reasonable, applying its supervision at an incompatible generation stage may still produce low-value or even competing gradients.

This issue resembles course scheduling in education. A student’s learning outcome depends not only on what is taught, but also on whether the course sequence matches the student’s current developmental stage.
The visualization in Fig.~\ref{fig:motivation} further shows how teacher-query and supervision-skipping timing affect overall performance. Teachers specializing in semantic structure achieve the best results when queried early in the generation stage. In contrast,  Optical Character Recognition (OCR) teachers focusing on textual content and teachers specializing in aesthetic quality yield greater supervisory benefits during the middle and late stages, respectively.

\section{Method}
This section presents PhaseOPD's overall framework, extending task-level teacher organization to generation-dynamics-aware teacher coordination. We first introduce an adaptive generation-stage identification module that analyzes the student model’s own generation trajectory to determine the generative responsibility of its current state. Based on the identified stage, PhaseOPD employs a dynamic teacher-routing mechanism that assigns teachers with different capabilities to the generation stages where their supervision is most effective, thereby reducing cross-stage supervision conflicts. Finally, we propose a dynamic rollout truncation mechanism to eliminate redundant computation.

\begin{figure*}[t]
    \centering
    \includegraphics[width=0.98\textwidth]{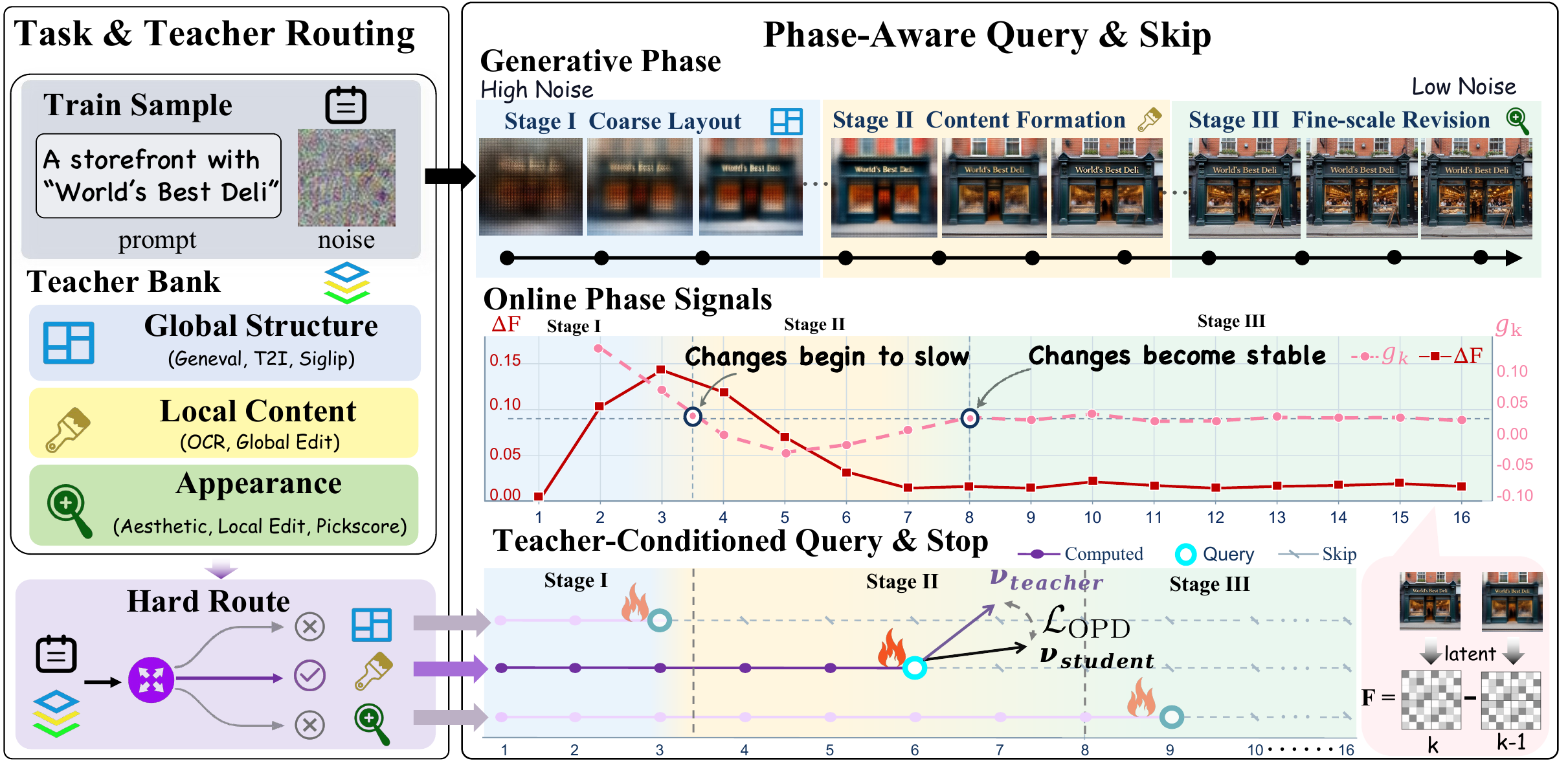}
    \caption{Overall framework of our method}
    \label{fig:picture2}
\end{figure*}

\subsection{Generative Models and On-Policy Distillation}
\paragraph{\textit{Flow-Matching based Generation}.}
For the denoising process,
$
    \mathbf{x}_t
    =
    (1-t)\mathbf{x}_0
    +
    t\mathbf{x}_1.
    \label{eq:flow_path}
$ where $t\in[0,1]$ denotes the continuous interpolation time, and $\mathbf{x}_t$ represents the noisy latent state at $t$.
$\mathbf{x}_0$ denotes clean data, while $\mathbf{x}_1$ represents pure noise.
The model learns a velocity field $\mathbf{v}_{\theta}(\mathbf{x}_t,t,\mathbf{c})$, where $\mathbf{c}$ denotes the prompt encoder. Using the predicted velocity, we obtain the current clean estimate:
\begin{equation}
    \hat{\mathbf{x}}_{0|t}
    =
    \mathbf{x}_t
    -
    t\,\mathbf{v}_{\theta}
    (\mathbf{x}_t,t,\mathbf{c}).
    \label{eq:clean_prediction}
\end{equation}
\paragraph{\textit{Reward-Based Policy Optimization}.}

Reinforcement learning (RL) provides a direct approach to optimizing task-specific performance during the post-training of generative models. Based on the continuous generation trajectory defined in the previous section, a generative model can be viewed as a parameterized policy $\pi_{\theta}$, where the complete generation process corresponds to a trajectory sampled from the policy $\tau \sim \pi_{\theta}$. Given the conditioning information $\mathbf{c}$, a reward function evaluates the quality of the final generated sample $r(\mathbf{x}_0,\mathbf{c})$. The RL objective is defined as maximizing the expected reward 
$
\mathcal{J}(\theta)
=
\mathbb{E}_{\mathbf{x}_0\sim\pi_{\theta}}
\left[
r(\mathbf{x}_0,\mathbf{c})
\right].
$
Using policy gradient methods, the model parameters are updated according to the reward feedback 
$
\nabla_{\theta}\mathcal{J}(\theta)
=
\mathbb{E}_{\tau\sim\pi_{\theta}}
\left[
\nabla_{\theta}
\log \pi_{\theta}(\tau)
r(\mathbf{x}_0,\mathbf{c})
\right].
$
Multi-objective generation tasks typically require the simultaneous optimization of multiple evaluation criteria 
$
\mathcal{R}
=
\{r_1,r_2,\cdots,r_K\},
$
where each reward targets a specific generative capability or preference.

\paragraph{\textit{Multi-Teacher On-Policy Distillation}.}
$
\{
\pi_{\phi_1},
\pi_{\phi_2},
\cdots,
\pi_{\phi_K}
\},
$
$
\tau \sim \pi_{\theta},
$

Multi-teacher On-Policy Distillation first trains multiple teacher policies for different objectives \citep{fang2026flowopd,li2026diffusionopd}
$
\{
\pi_{\phi_1},
\pi_{\phi_2},
\cdots,
\pi_{\phi_K}
\},
$
where $\pi_{\phi_k}$ denotes the teacher policy optimized for the $k^{th}$ objective. During distillation, the student model $\pi_{\theta}$ generates trajectories according to its own policy
$\tau \sim \pi_{\theta}$, and the corresponding teacher provides online supervision at the states the student visits. For the $k^{th}$ teacher, the distillation objective can be formulated as:
\begin{equation}
\mathcal{L}_{k}
=
\mathbb{E}_{\tau\sim\pi_{\theta}}
\left[
D
\left(
\pi_{\theta}(\cdot|\mathbf{x}_t),
\pi_{\phi_k}(\cdot|\mathbf{x}_t)
\right)
\right],
\end{equation}
where $D(\cdot,\cdot)$ measures the discrepancy between the student and teacher policies.


\subsection{Online Generative Phase Perception}
\label{sec:phase_perception}

Existing OPD methods select teachers primarily according to task categories. However, a teacher's supervisory value varies across states and generation stages. To allocate teacher supervision on demand following the inherent generation dynamics, PhaseOPD detects stage transitions during the student model’s on-policy rollout and adaptively determines when each teacher should provide supervision.


\paragraph{\textit{Spatial Dynamics Analysis}.}

Let $k$ denote a discrete sampling index along the student-generated trajectory, and let $t_k$ denote the corresponding continuous time. Using the clean latent predictions from Eq.~\ref{eq:clean_prediction}, we define the revision between adjacent sampling positions as:
\begin{equation}
\mathbf{d}_k
=
\hat{\mathbf{x}}_{0\mid t_k}
-
\hat{\mathbf{x}}_{0\mid t_{k-1}}.
\label{eq:latent_revision}
\end{equation}
To distinguish block-level structural changes from local spatial variations, let $\mathcal{P}_s(\cdot)$ denote a block-averaging projection at scale $s$, which first applies channel-wise average pooling with both kernel size and stride set to $s$, followed by nearest-neighbor upsampling to restore the original latent spatial dimensions. This process can be formulated as:
\begin{equation}
\begin{aligned}
&\bigl(\mathcal{P}_s(\mathbf{d}_k)\bigr)_{\ell,h,w}\\
&=
\frac{1}{s^2}
\sum_{a=0}^{s-1}
\sum_{b=0}^{s-1}
(\mathbf{d}_k)_{
\ell,\,
s\lfloor h/s\rfloor+a,\,
s\lfloor w/s\rfloor+b
}.
\end{aligned}
\label{eq:block_projection}
\end{equation}
Here, $\ell$ denotes the latent channel index; $h$ and $w$ denote the zero-based spatial row and column indices, respectively; $a$ and $b$ are within-block offsets; and $\lfloor\cdot\rfloor$ denotes the floor operation. This operation assigns the mean of each non-overlapping $s\times s$ block to all positions within that block and requires both latent's height and width to be divisible by $s$. We define the spatial-scale revision ratio as: 
\begin{equation}
F_k
=
\frac{
\left\|
\mathbf{d}_k-
\mathcal{P}_s(\mathbf{d}_k)
\right\|_2^2
}{
\left\|\mathbf{d}_k\right\|_2^2
+\varepsilon
}.
\label{eq:revision_ratio}
\end{equation}
Here, $\varepsilon$ is a numerical stability term, and we use $s=4$ in this work. A smaller $F_k$ indicates that the current revision is dominated by block-level, low-frequency structural changes. In contrast, a larger $F_k$ indicates a greater contribution of local spatial variations to the revision energy. This signal measures the spatial composition of the prediction revision, not image quality or overall revision magnitude. We further define the rate of change in the spatial-scale revision ratio between adjacent sampling positions as:

\begin{equation}
\Delta F_k
=
F_k-F_{k-1}.
\label{eq:revision_change}
\end{equation}

\paragraph{\textit{Adaptive Phase Boundary Detection}.}

According to the previously defined rate of change in the spatial-scale revision ratio, $\Delta F_k$, we further analyze how its slope evolves along the generation trajectory to identify dynamical boundaries between generation stages. Specifically, during denoising, $\Delta F_k$ initially increases, then decreases, and gradually stabilizes. We therefore define two stage boundaries as the first positions at which the slope approaches zero during the increasing and decreasing phases, respectively. First, we define the discrete slope of $\Delta F_k$ between adjacent sampling positions as 

\begin{equation}
g_k
=
\frac{\Delta F_k-\Delta F_{k-1}}{k-(k-1)}
=
\Delta F_k-\Delta F_{k-1}.
\label{eq:phase_slope}
\end{equation}
To prevent local fluctuations at individual sampling positions from obscuring the overall trend, we introduce a historical observation window of length 
$W$ and define its average slope as: 
\begin{equation}
\bar{g}_k^{(W)}
=
\frac{1}{W}
\sum_{j=k-W}^{k-1}g_j.
\label{eq:average_phase_slope}
\end{equation}
Here, $W$ denotes the length of the historical observation window, and $\bar{g}_k^{(W)}$ denotes the average slope over the $W$ sampling positions preceding the current position. The first stage boundary is defined as the first position at which the current slope approaches zero following a period of sustained growth in $\Delta F_k$ as:
\begin{equation}
k_{\mathrm{L}}
=
\min
\left\{
k
\;\middle|\;
\bar{g}_k^{(W)}>0,\;
|g_k|\to 0
\right\}.
\label{eq:first_phase_boundary}
\end{equation}
The second stage boundary is defined as the position at which the current slope again approaches zero following a period of sustained decline in $\Delta F_k$, as formulated:
\begin{equation}
k_{\mathrm{R}}
=
\min
\left\{
k>k_{\mathrm{L}}
\;\middle|\;
\bar{g}_k^{(W)}<0,\;
|g_k|\to 0
\right\}.
\label{eq:second_phase_boundary}
\end{equation}
Since the generation process consists of discrete sampling positions, we adopt a near-zero criterion rather than requiring the slope to be exactly zero. Based on the two dynamical boundaries defined above, we adaptively partition the student model’s generation trajectory into three consecutive stages as:
\begin{equation}
\mathrm{S}_k
=
\begin{cases}
\mathrm{I}, & k < k_{\mathrm{L}},\\
\mathrm{II}, & k_{\mathrm{L}} \leq k < k_{\mathrm{R}},\\
\mathrm{III}, & k \geq k_{\mathrm{R}}.
\end{cases}
\label{eq:generative_phase}
\end{equation}
Here, $\mathrm{S}_k$ denotes the generation stage at sampling position $k$. Stage $\mathrm{I}$ establishes global semantics and primary structures, Stage $\mathrm{II}$ forms local content, and Stage $\mathrm{III}$ refines visual aesthetics.

\subsection{Teacher-Conditioned Phase Routing}

Building on the three generation stages identified in Sec.\ref{sec:phase_perception} , we establish a correspondence between generation stages and supervision objectives, enabling each specialized teacher to contribute during the stage in which its supervision is most valuable.

\noindent\textbf{\textit{Stage I: Global Structure Formation}.}
Since the overall structure has not yet fully formed at this stage, local details lack a stable structural foundation. This stage is therefore better suited to supervision targeting global structural consistency, such as GenEval. We set the teacher query position to the sampling position immediately preceding the left boundary, namely $k_{\mathrm L}-1$. This choice is motivated by the observation that states too close to the pure-noise endpoint lack well-defined structure. Selecting the last sampling position before the left boundary thus allows structure-related supervision to be applied once structural information has begun to emerge, but before local details become dominant.

\noindent\textbf{\textit{Stage II: Local Content Generation}.}
In Stage II, the primary structure provides a relatively stable spatial foundation for local content, while local semantics and details continue to emerge and evolve. This stage is therefore well suited to specialized teachers that focus on local content accuracy, such as OCR teachers. We randomly sample a supervision point within Stage II to cover varying degrees of local content formation, avoid excessive reliance on a single generation position, and improve the adaptability of local content supervision to variations in states within this stage.

\noindent\textbf{\textit{Stage III: Visual Aesthetic Enhancement}.}
In the final stage of generation, the image approaches convergence, with the primary structure and local details remaining largely unchanged. This stability provides a foundation for aesthetic optimization, making this stage well suited to specialized aesthetic teachers, such as Aesthetic. In Stage III, we set the teacher query position to the first sampling position after the right boundary, namely $k_{\mathrm R}+1$. This choice lets aesthetic supervision act on a state where the primary structure and local content are already stable, while still leaving room for further refinement.

Once the supervision position for the current sample is determined, the teacher provides supervision only at the corresponding student-visited state, preserving the basic formulation of on-policy distillation while reducing cross-stage supervision mismatch and redundant trajectory computation. The unified query rule can be written as:
\begin{equation}
q_m=
\begin{cases}
k_{\mathrm L}-1,
& m\in\mathcal T_{\mathrm{str}},\\[3pt]
K,\quad K\sim\operatorname{Unif}
\bigl(\{k_{\mathrm L},\ldots,k_{\mathrm R}-1\}\bigr),
& m\in\mathcal T_{\mathrm{loc}},\\[3pt]
k_{\mathrm R}+1,
& m\in\mathcal T_{\mathrm{vis}}.
\end{cases}
\end{equation}
Here, $q_m$ denotes the query position for teacher 
$m$; $\mathcal{T}_{\mathrm{str}}$,  $\mathcal{T}_{\mathrm{loc}}$, and $\mathcal{T}_{\mathrm{vis}}$ denote the sets of teachers responsible for optimizing global structure, local content, and visual quality, respectively; and $K$ denotes a position randomly sampled from Stage II. We emphasize that GenEval, OCR, and Aesthetic serve as representative examples of these three teacher categories rather than a fixed teacher configuration. Other teachers with corresponding optimization capabilities are equally applicable.

\begin{CJK*}{UTF8}{gbsn}
\raggedbottom

\section{Experiments}
\label{sec:experiments}


This chapter evaluates \ourmethod{} around three questions: (1) whether staged querying can improve both generation quality and training cost with the primary teacher group; (2) whether these gains generalize across teacher combinations, student backbones, and task types; and (3) what roles adaptive triggering and early termination play, and why online signals can identify effective querying stages. We present the main results, generalization experiments, component ablations, and mechanistic analyses in that order. Detailed settings and additional results are provided in Appendix~\ref{sec:additional_results}.

\subsection{Experimental Setup}
\label{sec:experimental_setup}

\noindent \paragraph{Implementation Details \& Training Data}

We initialize an SD3.5-Medium student from the base model and train rank-32 LoRA adapters at $512\times512$ resolution using frozen Aesthetic, GenEval, and OCR teachers. Our default configuration uses 16-step on-policy rollouts, equal task sampling, 3,000 updates, and four H20 GPUs. Full data construction, guidance scales, and optimization settings are provided in Appendix~\ref{sec:app_detailed_setup}.

\noindent \paragraph{Metrics.}

We use AES on DrawBench, GenEval, and OCR as the primary metrics for aesthetics, compositional alignment, and text rendering. We additionally report HPSv3, PickScore, and SigLIP where applicable, and use ImageReward and VQAScore as held-out diagnostics.

\noindent \paragraph{Baselines.}

We compare against the SD3.5-M base model, three single-teacher models, DanceOPD, and DiffusionOPD. Our default model augments DanceOPD with phase-aware querying; we additionally apply the same mechanism to DiffusionOPD and FlowOPD and evaluate its transfer to the BAGEL and FLUX backbones.

\subsection{Main Results}
\label{sec:main_results}
\label{sec:qualitative_comparison}

\begin{table*}[t]
    \centering
    \caption{\textbf{Final generation quality, independent diagnostic metrics, and training cost.} Aesthetic, GenEval, and OCR are training objectives. ImageReward and VQAScore, shaded in light blue, are excluded from training and assess cross-metric generalization and potential reward hacking. The base model and individual teachers serve as capability references. Among multi-teacher methods, the best and second-best results are shown in bold and underlined, respectively.}
    \label{tab:main_results}
    \normalsize
    \setlength{\tabcolsep}{2.3pt}
    \begin{tabular}{lcccccc>{\columncolor[RGB]{239,246,250}}r>{\columncolor[RGB]{239,246,250}}r}
        \toprule
Method & Aesthetic $\uparrow$ & GenEval $\uparrow$ & OCR $\uparrow$ & Time (min) $\downarrow$ & Update (s) $\downarrow$ & GPU-h $\downarrow$ & ImageReward $\uparrow$ & VQAScore $\uparrow$ \\
        \midrule
        \multicolumn{9}{l}{\textit{Base model and single-teacher references}} \\
        \addlinespace[2pt]
        SD3.5-M Base & 5.3973 & 0.6620 & 0.5611 & -- & -- & -- & 0.9565 & 0.7416 \\
        Aesthetic Teacher & 6.2133 & 0.7246 & 0.5852 & -- & -- & -- & 1.5708 & 0.7840 \\
        GenEval Teacher & 5.2709 & 0.9336 & 0.6473 & -- & -- & -- & 0.8062 & 0.7225 \\
        OCR Teacher & 5.2665 & 0.6631 & 0.9428 & -- & -- & -- & 0.9243 & 0.7275 \\
        \midrule
        \multicolumn{9}{l}{\textit{Multi-teacher distillation methods}} \\
        \addlinespace[2pt]
        DanceOPD & 5.8185 & 0.8903 & 0.9289 & \underline{275.67} & \underline{5.5133} & \underline{18.3778}& 1.3536 & 0.7856 \\
        DiffusionOPD & \underline{6.1335} & \underline{0.9457} & \textbf{0.9516} & 967.73 & 38.7093 & 64.5156 & \underline{1.5432} & \underline{0.7907} \\
        \rowcolor{OursRow}
        \textbf{Ours} & \textbf{6.1705} & \textbf{0.9461} & \underline{0.9413} & \textbf{161.89} & \textbf{3.2379} & \textbf{10.7929} & \textbf{1.5586} & \textbf{0.7923} \\
        \bottomrule
    \end{tabular}
\end{table*}

\begin{figure}[!tbp]
    \centering
    \includegraphics[width=0.98\columnwidth]{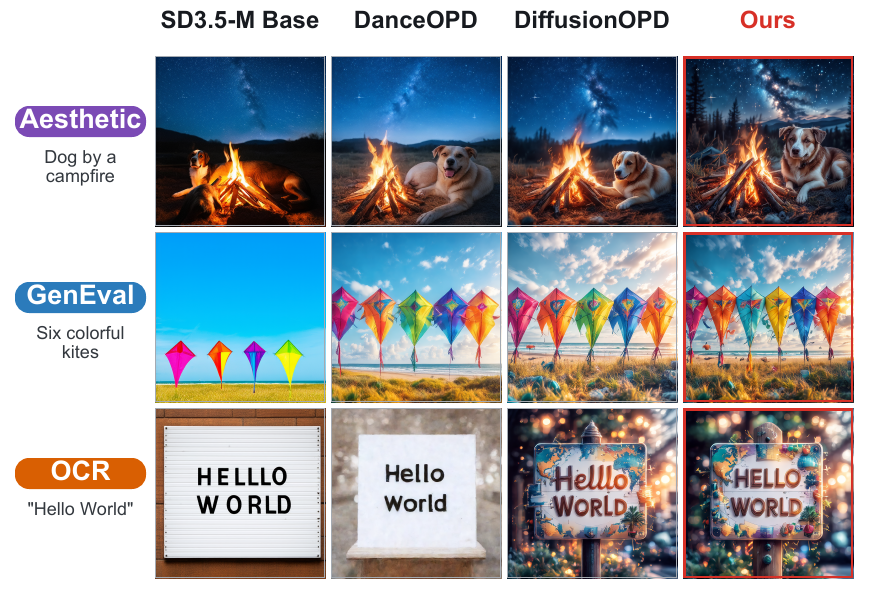}
    \caption{\textbf{Qualitative comparison with the primary teacher set.} Rows correspond to Aesthetic, GenEval, and OCR; columns show SD3.5-M Base, DanceOPD, DiffusionOPD, and Ours. All methods use archived, paired outputs generated with the same seed.}
    \label{fig:qualitative_primary}
\end{figure}

Tab.~\ref{tab:main_results} summarizes overall performance across multiple capabilities and training cost. Although DanceOPD and DiffusionOPD can integrate knowledge from multiple teachers, their trajectory-level supervision incurs substantial computational overhead. In contrast, our method achieves a better overall balance across Aesthetic, GenEval, and OCR while maintaining higher training efficiency. These results suggest that adaptively identifying and accommodating diffusion models' generation dynamics through targeted teacher supervision can improve multi-teacher knowledge fusion while avoiding unused rollout computation. Consistent improvements on independent diagnostic metrics—ImageReward and VQAScore in Tab.~\ref{tab:main_results}—further indicate that these gains generalize beyond the training objectives, rather than merely reflecting overfitting to teacher rewards. Fig.~\ref{fig:qualitative_primary} further presents representative results for the primary teacher group. We select Aesthetic, GenEval, and OCR prompts from a fixed sampling list. All methods within each row use the same prompt, initial noise, and inference configuration, so differences in visual quality, compositional generation, and text rendering directly reflect the three corresponding metrics in Tab.~\ref{tab:main_results}.

\begin{figure}[t]
    \centering
    \includegraphics[width=\columnwidth]{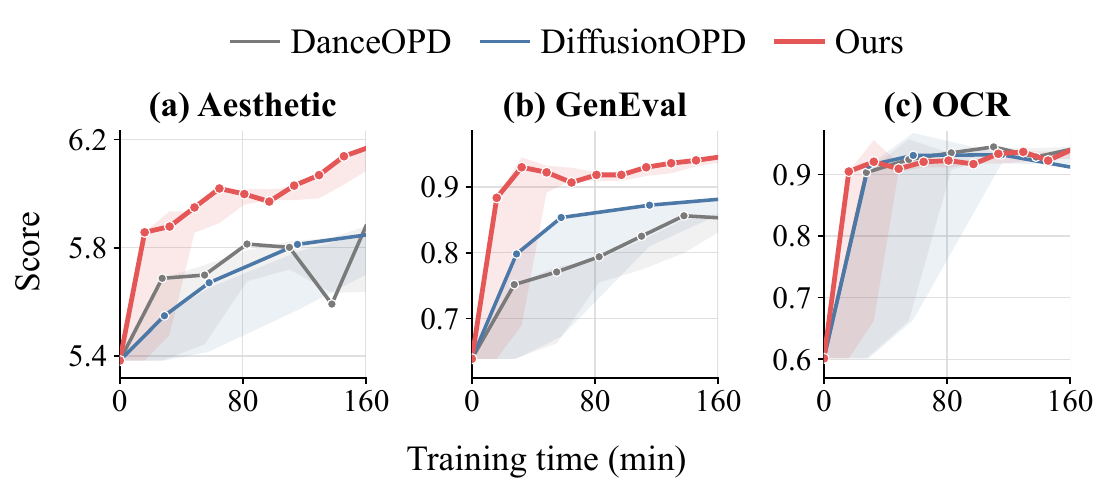}
    \caption{\textbf{Quality over the first 160 training minutes.} Panels show Aesthetic, GenEval, and OCR for DanceOPD, DiffusionOPD, and Ours. Time uses each method's matched mean update time measured with exclusive GPU access. }
    \label{fig:quality_efficiency}
\end{figure}

We further compare the cost required by different methods to reach the same visual metric scores. Fig.~\ref{fig:quality_efficiency} shows the learning curves of the three methods under the same wall-clock budget. Our method reaches competitive quality levels earlier and maintains stable performance across multiple capabilities throughout subsequent training, indicating that its advantages do not stem from a longer training budget or a fortuitously selected final checkpoint. Together with the final results in  Tab.~\ref{tab:main_results}, these findings show that our method consistently achieves a more favorable quality–time trade-off throughout training. The respective contributions of staged querying and early termination are further examined in the subsequent component ablations.

\begin{figure}[t]
    \centering
    \includegraphics[width=\columnwidth]{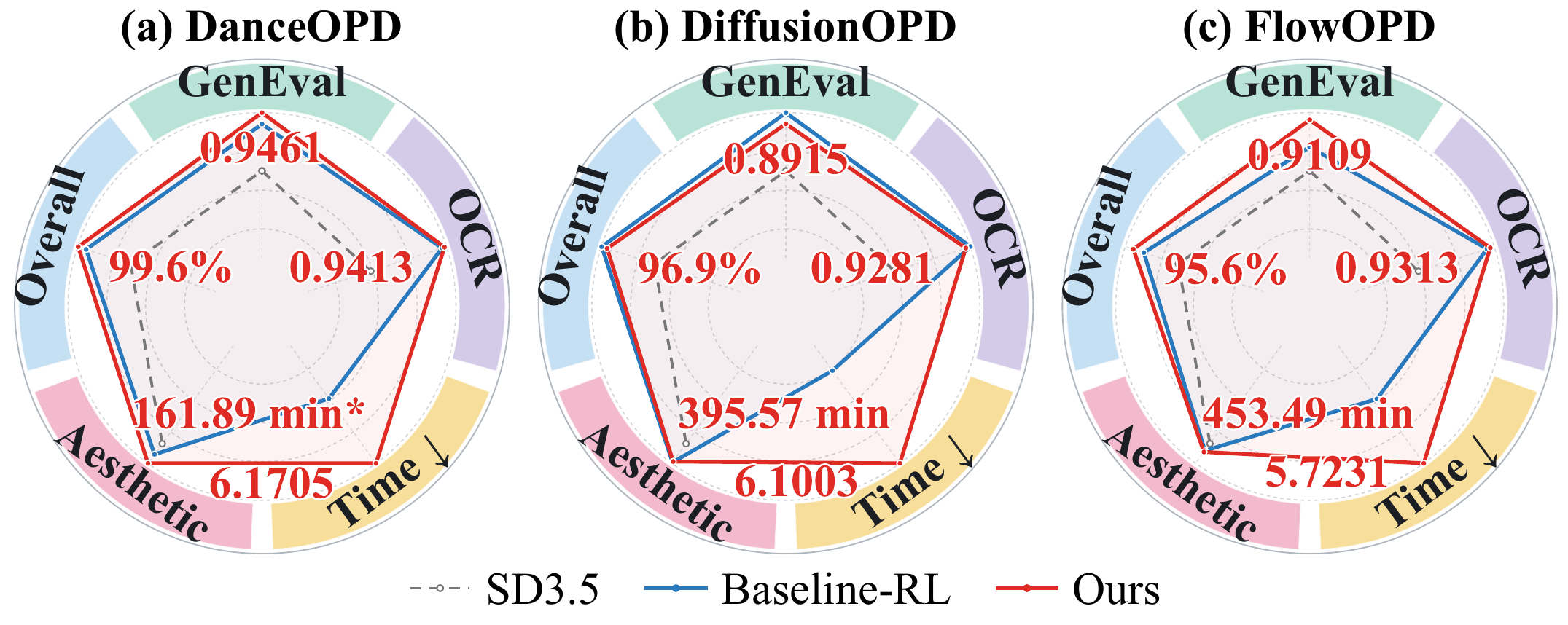}
    \caption{\textbf{Plug-and-play comparison across three OPD objectives.} Each panel compares SD3.5-M Base, the corresponding OPD method, and the method augmented with Ours. Quality metrics share scales across panels; Overall Quality is the geometric mean of the three normalized quality scores. Training efficiency uses inverse training time normalized within each panel and is omitted for the base model.}
    \label{fig:plugin_radar_three_host}
\end{figure}

Phase-aware querying can be directly integrated into DanceOPD, DiffusionOPD, and FlowOPD without modifying their original distillation objectives. As shown in Fig.~\ref{fig:plugin_radar_three_host}, incorporating our method improves both generation quality and training efficiency across all three underlying methods. These results indicate that PhaseOPD is not tied to a specific distillation loss or implementation and can serve as a general-purpose component for different OPD methods.

\subsection{Compatibility Experiments}
\label{sec:teacher_set_generalization}

\begin{table*}[t]
    \centering
    \caption{\textbf{Generation quality, cross-metric generalization, and training cost with alternative teachers.} HPSv3, SigLIP, and OCR are training objectives; ImageReward and VQAScore are held-out metrics for assessing generalization and potential reward hacking. Bold and underlined values indicate the best and second-best multi-teacher results, respectively. Base and single-teacher models serve as references. Ours offers a favorable quality--efficiency balance across teacher combinations.}
    \label{tab:second_teacher_main}
    \normalsize
    \setlength{\tabcolsep}{1.6pt}
    \begin{tabular}{lcccccc>{\columncolor[RGB]{239,246,250}}r>{\columncolor[RGB]{239,246,250}}r}
        \toprule
        Method & HPSv3 $\uparrow$ & SigLIP $\uparrow$ & OCR $\uparrow$ & Time (min) $\downarrow$ & Update (s) $\downarrow$ & GPU-h $\downarrow$ & ImageReward $\uparrow$ & VQAScore $\uparrow$ \\
        \midrule
        \multicolumn{9}{l}{\textit{Base model and single-teacher references}} \\
        \addlinespace[2pt]
        SD3.5-M Base & 2.613 & 0.1472 & 0.5611 & -- & -- & -- & 0.9565 & 0.7416 \\
        HPSv3 Teacher & 12.44 & 0.1361 & 0.5648 & -- & -- & -- & 1.309 & 0.7392 \\
        SigLIP Teacher & 7.091 & 0.1973 & 0.6083 & -- & -- & -- & 1.274 & 0.7904 \\
        OCR Teacher & 2.673 & 0.1507 & 0.9428 & -- & -- & -- & 0.9243 & 0.7275 \\
        \midrule
        \multicolumn{9}{l}{\textit{Multi-teacher distillation methods}} \\
        \addlinespace[2pt]
        DanceOPD & \underline{10.20} & \underline{0.1891} & \textbf{0.9418} & \underline{150.7} & \underline{3.014} & \underline{10.05} & \underline{1.299} & \underline{0.7635} \\
        DiffusionOPD & 9.964 & 0.1652 & \underline{0.9311} & 351.4 & 7.029 & 23.43 & 1.121 & 0.7438 \\
        \rowcolor{OursRow}
        \textbf{Ours} & \textbf{10.30} & \textbf{0.1938} & 0.9136 & \textbf{91.50} & \textbf{1.830} & \textbf{6.100} & \textbf{1.323} & \textbf{0.7665} \\
        \bottomrule
    \end{tabular}
\end{table*}

\begin{figure}[!tbp]
    \centering
    \includegraphics[width=\columnwidth]{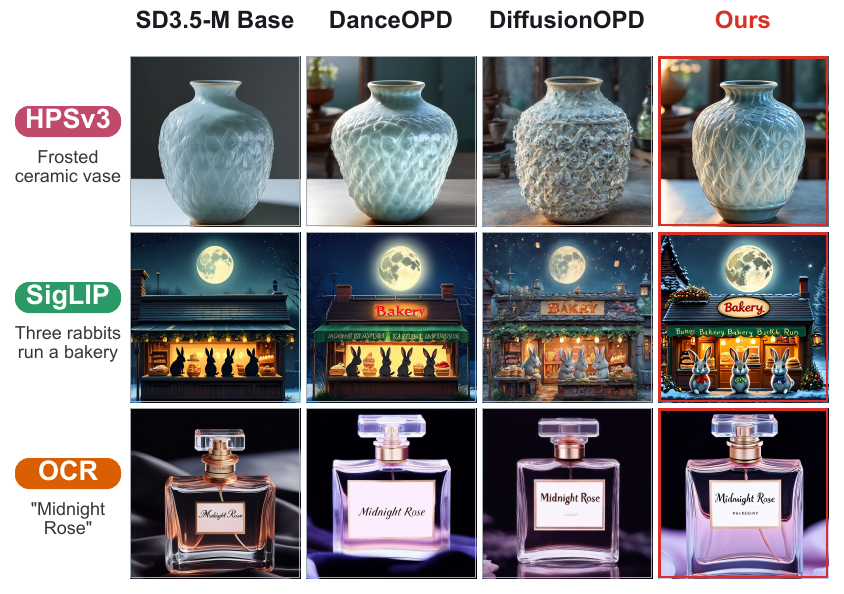}
    \caption{\textbf{Qualitative comparison with the alternative teacher set.} Rows correspond to HPSv3, SigLIP, and OCR; columns show SD3.5-M Base, DanceOPD, DiffusionOPD, and Ours.}
    \label{fig:qualitative_alternate}
\end{figure}

\begin{figure}[t]
    \centering
    \includegraphics[width=\columnwidth]{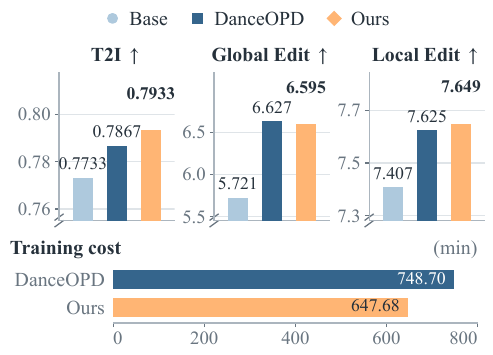}
    \caption{\textbf{Cross-backbone and cross-task generalization on BAGEL.} This setting changes both the student backbone and teacher tasks, covering text-to-image generation, global editing, and local editing.}
    \label{fig:bagel_generalization}
\end{figure}

\newsavebox{\fluxgeneralizationbox}
\begin{lrbox}{\fluxgeneralizationbox}
\begin{minipage}[t]{0.65\textwidth}
    \vspace{0pt}
    \centering
    \captionsetup{type=table,hypcap=false,font=footnotesize,aboveskip=0pt,belowskip=4pt}
    \caption{\textbf{Cross-backbone generalization on FLUX.} FLUX.1-dev Base serves as the pretrained reference. We compare DanceOPD and Ours on PickScore, GenEval, OCR, and training cost.}
    \label{tab:flux_generalization}
    \fontsize{8}{10}\selectfont
    \renewcommand{\arraystretch}{1.0}
    \setlength{\aboverulesep}{1.5pt}
    \setlength{\belowrulesep}{1.5pt}
    \setlength{\tabcolsep}{2.4pt}
    \begin{tabular*}{\linewidth}{@{\extracolsep{\fill}}lcccccc@{}}
        \toprule
        Method & PickScore $\uparrow$ & GenEval $\uparrow$ & OCR $\uparrow$ & \shortstack{Time\\(min)$\downarrow$} & \shortstack{Update\\(s)$\downarrow$} & GPU-h $\downarrow$ \\
        \midrule
        FLUX.1-dev Base & 0.8745 & 0.6569 & 0.56540 & -- & -- & -- \\
        \midrule
        DanceOPD & 0.9284 & 0.8271 & 0.8842 & 804.88 & 16.091 & 53.658 \\
        \rowcolor{OursRow}
        \textbf{Ours} & \textbf{0.9292} & \textbf{0.82858} & \textbf{0.9478} & \textbf{376.00} & \textbf{7.510} & \textbf{25.066} \\
        \bottomrule
    \end{tabular*}
\end{minipage}
\end{lrbox}

\begin{figure}[t]
    \centering
    \includegraphics[width=\columnwidth]{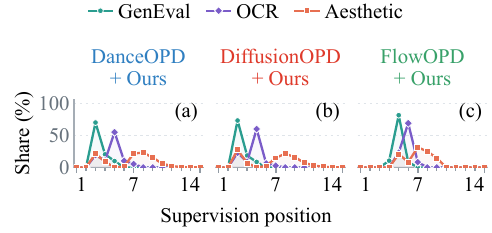}
    \caption{\textbf{Teacher-specific query distributions across three OPD hosts.} Panels correspond to Ours, DiffusionOPD + Ours, and FlowOPD + Ours, each showing GenEval, OCR, and Aesthetic teachers. The horizontal axis indicates query position; the vertical axis shows the percentage of trajectories normalized separately for each teacher.}
    \label{fig:three_host_query_distribution}
\end{figure}

\begin{figure}[t]
    \centering
    \includegraphics[width=\columnwidth]{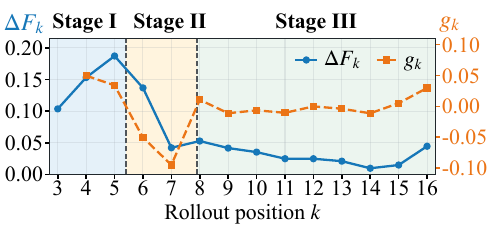}
    \caption{\textbf{Mean dynamics of online detection signals.} The blue curve shows the change in the spatial-scale revision ratio, $\Delta F_k$, and the orange curve shows its first-order difference $g_k$. Background colors and vertical dashed lines indicate three generation stages identified by near-zero slopes after sustained increases and decreases in $\Delta F_k$.}
    \label{fig:mean_dynamics}
\end{figure}

\begin{figure*}[!t]
    \centering
    \usebox{\fluxgeneralizationbox}\hfill
    \begin{minipage}[t]{0.32\textwidth}
        \vspace{0pt}
        \centering
        \includegraphics[width=\linewidth]{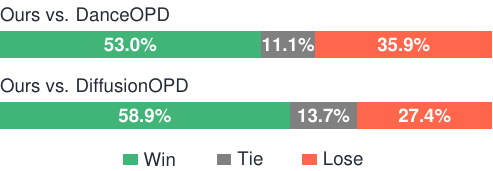}
        \caption{Double-blind user preferences for Ours versus multi-teacher baselines.}
        \label{fig:user_study}
    \end{minipage}
\end{figure*}

To examine whether our method depends on the default teacher combination, we replace the Aesthetic and GenEval teachers with HPSv3 and SigLIP while retaining the OCR teacher. Tab.~\ref{tab:second_teacher_main} shows that the single-teacher models still exhibit pronounced task-specific biases, whereas our method achieves the best HPSv3, SigLIP, and independent diagnostic scores among multi-teacher methods while remaining competitive on OCR. These results indicate that phase-aware querying can adapt to different teachers' capability profiles and still produce a well-balanced, multi-capability student under different teacher combinations, rather than being effective only for the default teacher group.

Fig.~\ref{fig:qualitative_alternate} further presents paired generation results across the three capabilities. The HPSv3 examples emphasize overall visual quality, the SigLIP examples assess image–text semantic alignment and spatial consistency, and the OCR examples focus on the accuracy and legibility of the specified text. Our method maintains consistent performance across different prompt types, providing qualitative support for the quantitative results obtained with the alternative teacher group.

\subsection{Generalization Experiments}
\label{sec:zimage_task_generalization}
\label{sec:flux_backbone_generalization}

We evaluate the cross-backbone generalization of our method in two complementary settings. The BAGEL experiments change both the generative backbone and the task formulation, extending text-to-image generation to global and local editing. As shown in Fig.~\ref{fig:bagel_generalization}, our method improves text-to-image generation and local editing, remains competitive on global editing, and reduces overall training cost. The FLUX experiments retain the multi-teacher text-to-image generation setting and replace only the underlying generative model. Tab.~\ref{tab:flux_generalization} shows that our method outperforms DanceOPD across all generation quality metrics while achieving higher training efficiency. Together, these experiments demonstrate that phase-aware querying is not tied to the specific generation process of SD3.5-M but can generalize across model architectures and task formulations, consistently improving the quality–cost trade-off.

\subsection{Ablation Studies}
\label{sec:ablation}

\begin{table*}[t]
    \centering
    \caption{\textbf{Effects of individual components on final quality and training efficiency.} AT denotes adaptive triggering, and ES denotes early skipping. The best and second-best results for each metric are shown in bold and underlined, respectively.}
    \label{tab:component_ablation}
    \small
    \setlength{\tabcolsep}{4.0pt}
    \begin{tabular}{lcccccccc}
        \toprule
        Variant & AT & ES & Aesthetic $\uparrow$ & GenEval $\uparrow$ & OCR $\uparrow$ & Time (min)$\downarrow$ & Update (s)$\downarrow$ & GPU-h $\downarrow$ \\
        \midrule
        Ours (w/o AT and ES) & $\times$ & $\times$ & 5.9044 & \textbf{0.9496} & 0.8626 & 275.333 & 5.5067 & 18.356 \\
        Ours (w/o AT) & $\times$ & $\checkmark$ & 5.8073 & 0.8892 & \underline{0.9305} & \underline{241.78} & \underline{4.836} & \underline{16.119} \\
        Ours (w/o ES) & $\checkmark$ & $\times$ & \underline{6.1642} & 0.9341 & 0.9275 & 277.667 & 5.5533 & 18.511 \\
        \rowcolor{OursRow}
        \textbf{Full Ours} & $\checkmark$ & $\checkmark$ & \textbf{6.1705} & \underline{0.9461} & \textbf{0.9413} & \textbf{161.893} & \textbf{3.2379} & \textbf{10.793} \\
        \bottomrule
    \end{tabular}
\end{table*}

Our method comprises adaptive phase querying (AT) and early skipping (ES). To examine their respective roles, we remove AT, ES, and both components in Tab.~\ref{tab:component_ablation}. Removing ES yields Aesthetic performance close to that of the full model, with slight decreases in GenEval and OCR and a substantial increase in training time. This suggests that ES effectively reduces rollout computation while preserving generation quality. Removing only AT lowers all three quality metrics and increases training time, indicating that adaptive phase querying helps balance effective supervision with computational efficiency. When we remove both AT and ES, GenEval improves, but Aesthetic and OCR decline substantially, revealing trade-offs among capabilities. The full model achieves the best Aesthetic and OCR scores and the second-best GenEval score, while offering the highest training efficiency. These results support the complementary roles of AT and ES: AT selects supervision points according to the generation phase, while ES eliminates subsequent computation once the necessary queries have been completed. Together, they achieve a better balance between generation quality and training efficiency.

\subsection{Mechanism Analysis}
\label{sec:routing_analysis}

Controlled single-teacher experiments on phase sensitivity are provided in Appendix~\ref{sec:query_audit}.
To analyze the querying preferences of different teachers during generation, we summarize the query positions selected by our method under three OPD training objectives in Fig.~\ref{fig:three_host_query_distribution}. GenEval queries are concentrated primarily in the early stages, OCR queries favor the middle stages, and Aesthetic queries span both early and late stages. Although the precise peaks vary with the underlying training objective, the overall phase preferences of the three teachers remain consistent. These results indicate that different teachers supervise at different generation stages, while our method adapts query positions to the student's actual trajectory rather than relying on fixed timesteps.

\noindent \paragraph{Student dynamics recover this ordering online} 
Fig.~\ref{fig:mean_dynamics} shows the two online signals used by the adaptive trigger: the change in the spatial-scale revision ratio, $\Delta F_k=F_k-F_{k-1}$, and its first-order difference, $g_k=\Delta F_k-\Delta F_{k-1}$. As the rollout progresses, $\Delta F_k$ rises rapidly in the early stages and reaches a peak, then declines substantially before settling into small, stable fluctuations in the later stages. The detector identifies stage boundaries when $g_k$ approaches zero after sustained increases and decreases in $\Delta F_k$, respectively, following the criterion in Sec.~\ref{sec:phase_perception}. These transitions separate the early rise, intermediate decline, and later stabilization of the spatial-scale revision signal. Because both signals are computed solely from current and past student states, phase boundaries can be identified causally online without access to future rollout states.

\subsection{User Preference Study}
\label{sec:user_study}

To evaluate the generated results from a human perceptual perspective, we conduct a double-blind user preference study comparing our method against DanceOPD and DiffusionOPD in pairwise comparisons. Each pair of candidate images is generated using the same text prompt, random seed, and inference configuration and presented in a randomized order. Participants are unaware of which method produced each image and are asked to select the better result based on overall visual quality, text alignment, and content completeness, or to indicate a tie. As shown in Fig.~\ref{fig:user_study}, our method receives higher user preference rates against both multi-teacher distillation baselines. These findings are consistent with the automatic evaluations, indicating that the improvements from phase-aware querying translate into perceptible gains in generation quality rather than merely numerical improvements on specific evaluation metrics.

\end{CJK*}

\section{Conclusion}
\label{sec:conclusion}
We presented PhaseOPD, which connects teacher supervision timing with early termination of student rollouts. Causal signals from predicted-clean latent revisions identify teacher-specific query states, allowing single-query distillation to proceed without generating the remaining trajectory. Experiments show fewer rollout steps, lower training cost, the strongest GenEval score in our comparison, and faster first-hour gains on GenEval and OCR. The results support jointly designing \emph{where to learn} and \emph{how far to roll out} as a practical route to efficient multi-capability post-training.

\clearpage
{
    \small
    \bibliographystyle{ieeenat_fullname}
    \bibliography{main}
}

\clearpage
\appendix
\begin{CJK*}{UTF8}{gbsn}

\section{Related Work}
\label{sec:related_work}

\subsection{Post-training generative models.}
Diffusion models \citep{ho2020ddpm} and flow matching \citep{lipman2023flow} learn iterative transformations between noise and data; rectified-flow transformers scale this approach to high-resolution text-to-image synthesis \citep{esser2024scaling}. Post-training methods align these models with downstream objectives. DDPO treats denoising as a sequential decision process and applies policy gradients to image rewards \citep{black2024ddpo}; Diffusion-DPO instead learns from pairwise preferences \citep{wallace2024diffusiondpo}. Flow-GRPO extends online policy optimization to flow matching and reduces the number of training denoising steps \citep{liu2025flowgrpo}. PhaseOPD integrates expert capabilities and reduces rollout computation by identifying supervision states online.

\subsection{On-policy capability transfer.}
Training on student-generated sequences is a central principle of on-policy language-model distillation \citep{agarwal2024gkd}. DiffusionOPD transfers task-specific teachers along student trajectories using a continuous-state distillation formulation \citep{li2026diffusionopd}. Flow-OPD combines specialized teachers, task routing, and dense trajectory supervision for flow models \citep{fang2026flowopd}. DanceOPD routes each sample to one generative field and uses velocity regression at a low-noise student state \citep{zhou2026danceopd}. We retain task-specific teachers and single-query regression. Our contribution concerns how student dynamics determine the query state and the minimum rollout prefix needed to select it; we do not claim a new regression loss or single-query distillation principle.

\subsection{Stage specialization and efficient sampling.}
eDiff-I assigns separately trained denoisers to different synthesis stages \citep{balaji2022ediffi}, demonstrating the usefulness of time-dependent specialization. PhaseOPD instead schedules supervision from capability-specific teachers using the evolving student state, producing one student for inference. Progressive distillation reduces the number of steps required to generate a completed sample \citep{salimans2022progressive}. Our early stopping serves a different purpose: it removes the unneeded suffix of a \emph{training} trajectory once its supervision state is known. It leaves the inference procedure unchanged and can be studied independently of methods that accelerate completed-image sampling.

\subsection{On-Policy Distillation}
\label{sec:opd_related_work}
On-policy distillation supervises students on self-generated states, reducing training--generation mismatch \citep{agarwal2024gkd}. For visual generation, Flow-OPD combines task routing with dense trajectory supervision \citep{fang2026flowopd}, while DiffusionOPD derives a continuous-state distillation objective \citep{li2026diffusionopd}. Flow3D-OPD extends multi-teacher transfer to 3D geometry \citep{ning2026flow3dopd}, and DanceOPD uses one low-noise velocity query per sample \citep{zhou2026danceopd}. These methods leave a complementary timing question: which state best supports each teacher's objective, and how much rollout is needed to select it? PhaseOPD uses student dynamics to determine teacher-specific query positions and skips the remaining rollout once all query decisions are finalized.

\section{Further Results}
\label{sec:additional_results}

\subsection{Detailed Setup}
\label{sec:app_detailed_setup}

\noindent\paragraph{Implementation Details \& Training Data.}
We use SD3.5-Medium as the base model for multi-teacher distillation at a resolution of $512\times512$. The student is initialized from the base model and trained using LoRA with rank $r=32$ and scaling factor $\alpha=64$. We use frozen task-specific teachers for Aesthetic, GenEval, and OCR, training the student to match their velocity predictions at selected generation states rather than directly optimizing reward scores. For Aesthetic, we use text prompts from Pick-a-Pic; for GenEval and OCR, we follow the training splits used in FlowGRPO. After deduplicating prompts, removing overlaps with the designated test splits, and setting aside fixed validation sets, we retain 25,106, 32,943, and 19,392 training prompts for the three tasks, respectively. Training uses 16-step online student rollouts, with a classifier-free guidance (CFG) scale of 4.5 for the student and 4.5, 1.0, and 4.5 for the three teachers, respectively. Tasks are sampled with equal probability, and all distillation loss weights are set to 1. The main experiments are run on four H20 GPUs for 3,000 updates with a learning rate of $10^{-4}$.

\noindent\paragraph{Metrics.}
We evaluate image aesthetic quality using Aesthetic Score (AES) on DrawBench and assess semantic alignment and text-generation accuracy using GenEval and OCR, respectively. We additionally report HPSv3, PickScore, SigLIP, ImageReward, and VQAScore as auxiliary evaluation metrics.

\noindent\paragraph{Baselines.}
We compare the base model, three single-teacher models, and representative multi-teacher online policy distillation methods, including DanceOPD and DiffusionOPD. The base and single-teacher models serve as references for individual capabilities, while the multi-teacher methods evaluate different knowledge-fusion strategies. By default, we use DanceOPD as the underlying distillation objective and refer to its variant augmented with phase-aware querying as \emph{Ours}. When the same strategy is applied to other objectives, we use the names ``DiffusionOPD + Ours'' and ``FlowOPD + Ours.'' In addition to the main experiments on SD3.5-M, we compare DanceOPD with Ours on the BAGEL and FLUX backbones to separate the effects of teacher combinations, generative backbones, and task types.

\subsection{Time to Matched Quality}
\label{sec:app_matched_quality}

\begin{figure}[!htbp]
    \centering
    \captionsetup{skip=3pt}
    \includegraphics[width=\columnwidth]{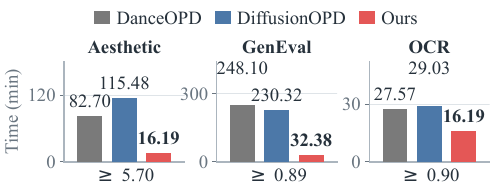}
    \caption{\textbf{Training time to reach quality targets:} Aesthetic $\geq 5.70$, GenEval $\geq 0.89$, and OCR $\geq 0.90$.}
    \label{fig:app_equal_quality_time_bars}
\end{figure}

To examine whether the training-time advantage persists at comparable quality, we measure the wall-clock time required to reach a common threshold for each task. Fig.~\ref{fig:app_equal_quality_time_bars} reports results for Aesthetic $\geq 5.70$, GenEval $\geq 0.89$, and OCR $\geq 0.90$. Ours reaches these targets in 16.19, 32.38, and 16.19 minutes, respectively, earlier than both DanceOPD and DiffusionOPD. These gains complement the final-checkpoint comparisons, showing that phase-aware querying and early skipping speed up capability acquisition across all three objectives.

\FloatBarrier
\subsection{Independent Diagnostic Evaluation}
\label{sec:reward_diagnostics}

We use independent diagnostic metrics to assess whether improvements extend beyond the rewards used to train the teachers. The seven models in each comparison are evaluated on GenAI-Bench-1600 with shared prompts, seeds, and inference settings: $512\times512$ resolution, 40 sampling steps, and CFG 4.5. ImageReward uses ImageReward-v1.0, while VQAScore uses CLIP-FlanT5-XXL through \texttt{t2v-metrics==3.0}. The results in Tabs.~\ref{tab:main_results} and~\ref{tab:second_teacher_main} show that Ours improves both diagnostic metrics over the multi-teacher baselines under both teacher combinations. This consistency supports transfer beyond the teacher-specific training objectives.

\FloatBarrier
\subsection{Query Intervals and Teacher Routing}
\label{sec:query_audit}

\begin{figure*}[!t]
    \centering
    \includegraphics[width=0.96\textwidth]{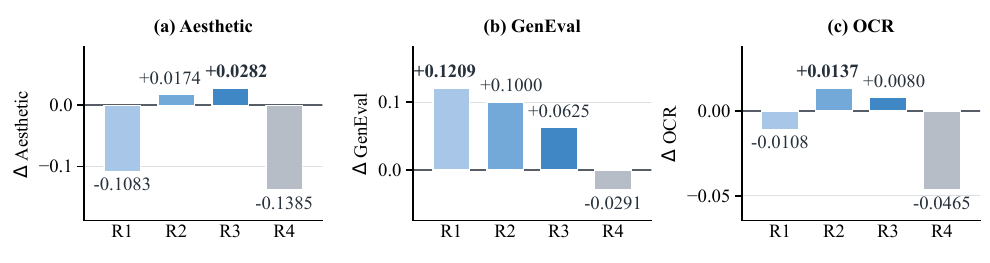}
    \caption{\textbf{Stage sensitivity of teacher supervision.} The three panels report Aesthetic, GenEval, and OCR gains from controlled single-teacher experiments. Each bar shows the change relative to the Beta query strategy when queries are restricted to the corresponding generation interval.}
    \label{fig:teacher_stage_sensitivity}
\end{figure*}

\noindent \paragraph{Phase Sensitivity of Teacher Supervision.}
Different teacher models exhibit different sensitivities to the querying phase. We conduct separate training runs with the Aesthetic, GenEval, and OCR teachers, restricting queries to R1 (steps 1--4), R2 (5--8), R3 (9--12), or R4 (13--16). Across these experiments, only the teacher query interval is varied. Fig.~\ref{fig:teacher_stage_sensitivity} reports changes in the corresponding target metric relative to the Beta querying strategy, reflecting each teacher’s individual phase preference rather than the joint performance of a multi-teacher model.

The GenEval teacher exhibits a clear preference for early-stage supervision, with its largest relative gain occurring in R1. Effective supervision from the OCR teacher concentrates in the middle stages, with the largest relative gain in R2 and the second-largest in R3. The Aesthetic teacher achieves its highest relative gain in R3, followed by R2. R4 underperforms the Beta strategy for all three teachers, indicating that states closer to the trajectory endpoint do not consistently provide more effective teacher supervision. These distinct temporal response profiles provide controlled experimental evidence for determining query timing based on the teacher type and the student’s current state.

\noindent\paragraph{Absolute Scores for Fixed Query Intervals.}
Tab.~\ref{tab:app_fixed_stage} provides the absolute scores underlying Fig.~\ref{fig:teacher_stage_sensitivity}. Each metric column comes from separate single-teacher training runs, with queries restricted to R1 (positions 1--4), R2 (5--8), R3 (9--12), or R4 (13--16); Beta is the reference query strategy. GenEval performs best in R1, OCR in R2, and Aesthetic in R3, while R4 falls below Beta for all three objectives. These results reinforce the distinct temporal preferences of the teachers and explain why a shared late-stage query rule can limit capability transfer.

\begin{table}[H]
    \centering
    \caption{Single-teacher scores for fixed query intervals.}
    \label{tab:app_fixed_stage}
    \small
    \setlength{\tabcolsep}{8pt}
    \begin{tabular}{lrrr}
        \toprule
        Query interval & Aesthetic & GenEval & OCR \\
        \midrule
        R1 & 6.3258 & \textbf{0.8042} & 0.8926 \\
        R2 & 6.4515 & 0.7833 & \textbf{0.9171} \\
        R3 & \textbf{6.4623} & 0.7458 & 0.9114 \\
        R4 & 6.2956 & 0.6542 & 0.8569 \\
        Beta & 6.4341 & 0.6833 & 0.9034 \\
        \bottomrule
    \end{tabular}
\end{table}

\noindent\paragraph{OCR Stage Policy and Teacher Routing.}
We further examine how the OCR query interval and task-routing distribution affect the final capability balance. Tab.~\ref{tab:app_stage_mixture} reports results after 3,000 updates. The stage-policy comparison fixes balanced routing and varies the OCR stage policy; the routing comparison fixes the Stage-1/2 mixture and varies the routing distribution. Both Ours rows reproduce the full configuration in Tab.~\ref{tab:main_results}. The Stage-1/2 mixture exceeds Stage-2-only and Stage-2/3 supervision on all three metrics, while balanced routing also outperforms OCR-biased routing across the reported scores. These comparisons show that allocating supervision across both trajectory positions and teacher tasks affects the joint capability balance.

\begin{table}[t]
    \centering
    \caption{OCR stage-policy and teacher-routing results at 3,000 updates. Both Ours rows reproduce Table~\ref{tab:main_results}.}
    \label{tab:app_stage_mixture}
    \fontsize{9}{11}\selectfont
    \renewcommand{\arraystretch}{1.2}
    \setlength{\tabcolsep}{3pt}
    \begin{tabular*}{\columnwidth}{@{\extracolsep{\fill}}lrrr@{}}
        \toprule
        Configuration & Aesthetic $\uparrow$ & GenEval $\uparrow$ & OCR $\uparrow$ \\
        \midrule
        \multicolumn{4}{@{}l}{\textit{OCR stage policy}} \\
        Stage-2 only & 5.9199 & 0.9379 & 0.9139 \\
        Stage-1/2 mixture \textbf{(Ours)} & 6.1705 & 0.9461 & 0.9413 \\
        Stage-2/3 mixture & 5.9154 & 0.9339 & 0.9139 \\
        \midrule
        \multicolumn{4}{@{}l}{\textit{Teacher routing policy}} \\
        Balanced \textbf{(Ours)} & 6.1705 & 0.9461 & 0.9413 \\
        OCR-biased & 5.8790 & 0.9456 & 0.9203 \\
        \bottomrule
    \end{tabular*}
\end{table}

\FloatBarrier
\raggedbottom
\subsection{Alternative-Teacher Learning Curves}
\label{sec:app_generalization}

\begin{figure}[H]
    \centering
    \includegraphics[width=\columnwidth]{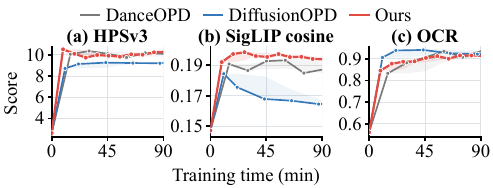}
    \caption{\textbf{Time--quality trajectories with the alternative teacher set under a 90-minute budget.} The panels report HPSv3, SigLIP cosine, and OCR. All methods start from the same SD3.5-M Base scores.}
    \label{fig:app_alt_trajectory}
\end{figure}

To examine learning efficiency under a different teacher combination, we replace the primary teachers with HPSv3, SigLIP, and OCR while retaining the student backbone, optimizer, and evaluation interval. Fig.~\ref{fig:app_alt_trajectory} compares DanceOPD, DiffusionOPD, and Ours over a common 90-minute budget. All curves start from the same SD3.5-M Base evaluation point; checkpoint times are converted from update counts using the measured training times in Tab.~\ref{tab:second_teacher_main}, and each trajectory is truncated at the budget boundary. The panels track perceptual preference, semantic alignment, and text accuracy over time, complementing the final-checkpoint scores.

\FloatBarrier
\subsection{Sensitivity to Training Guidance}
\label{sec:app_cfg_sensitivity}

We compare training CFG scales of 1.0 and 4.5 while fixing evaluation CFG at 4.5. The student and the Aesthetic and OCR teachers use the listed training scale, while the GenEval teacher remains at CFG 1.0. As shown in Tab.~\ref{tab:app_diffusion_cfg}, DanceOPD + Ours achieves higher scores on all three metrics at CFG 4.5, whereas CFG 1.0 reduces training time from 161.89 to 81.39 minutes. For DiffusionOPD + Ours, CFG 1.0 reduces time from 395.57 to 205.94 minutes and improves GenEval, while CFG 4.5 gives higher Aesthetic and OCR scores. Thus, the guidance scale offers a further quality--cost trade-off within both distillation frameworks.

\begingroup
\setlength{\textfloatsep}{4pt}
\setlength{\floatsep}{6pt}
\setlength{\intextsep}{2pt}
\captionsetup[table]{skip=2pt}
\renewcommand{\arraystretch}{0.9}
\begin{table}[H]
    \centering
    \caption{Quality and training cost under different training CFG settings. Evaluation CFG is fixed at 4.5.}
    \label{tab:app_diffusion_cfg}
    \fontsize{9}{11}\selectfont
    \renewcommand{\arraystretch}{1.15}
    \setlength{\tabcolsep}{3pt}
    \begin{tabular*}{\columnwidth}{@{\extracolsep{\fill}}crrrrrr@{}}
        \toprule
        CFG & Step & \shortstack{Time\\(min)} & GPU-h & Aesthetic & GenEval & OCR \\
        \midrule
        \multicolumn{7}{@{}l}{\textit{DanceOPD + Ours}} \\
        4.5 & 3000 & 161.89 & 10.7929 & \textbf{6.1705} & \textbf{0.9461} & \textbf{0.9413} \\
        1.0 & 3000 & \textbf{81.39} & \textbf{5.640} & 5.6135 & 0.9396 & 0.9129 \\
        \midrule
        \multicolumn{7}{@{}l}{\textit{DiffusionOPD + Ours}} \\
        4.5 & 1500 & 395.57 & 26.4101 & \textbf{6.1003} & 0.8915 & \textbf{0.9281} \\
        1.0 & 1500 & \textbf{205.94} & \textbf{13.7814} & 5.8450 & \textbf{0.9301} & 0.9239 \\
        \bottomrule
    \end{tabular*}
\end{table}

\FloatBarrier
\endgroup

\FloatBarrier
\onecolumn
\clearpage
\raggedbottom
\setlength{\intextsep}{2pt}
\captionsetup{font=normalsize,skip=2pt}
\interlinepenalty=10000
\subsection{Additional Qualitative Comparisons}
\label{sec:additional_qualitative}

We provide additional examples to examine how the transferred capabilities appear in generated images. The following comparisons cover text accuracy, visual presentation, semantic constraints, alternative teachers, and different OPD hosts, extending the quantitative evaluations with task-specific visual evidence.

\subsubsection{Text Accuracy}
\label{sec:app_text_accuracy}
\label{sec:app_text_visual}

\begin{figure}[H]
    \centering
    \includegraphics[width=0.86\textwidth]{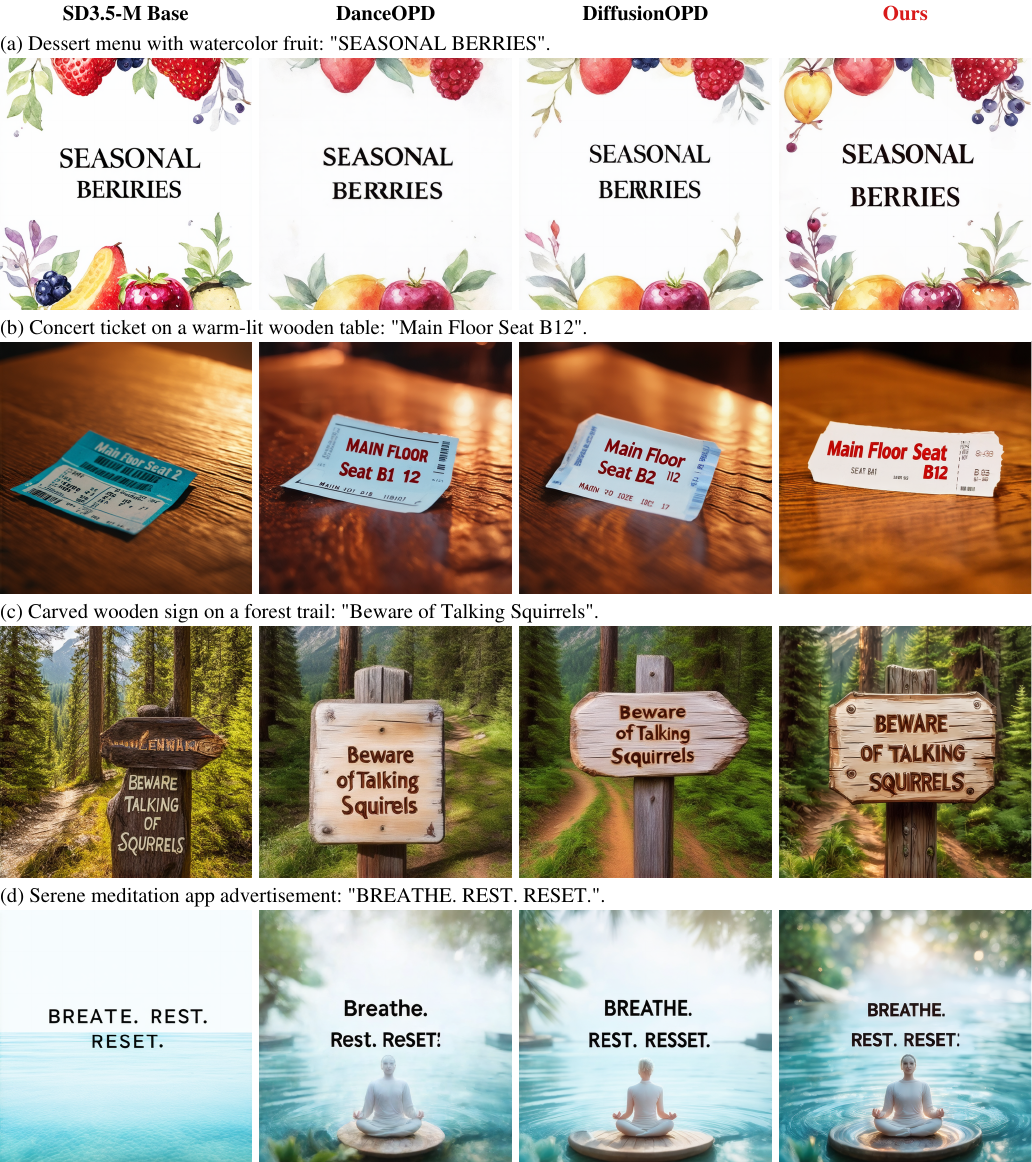}
    \caption{Text accuracy. PhaseOPD improves target-text reproduction; final punctuation in (d) remains imperfect.}
    \label{fig:app_qual_text}
\end{figure}

Fig.~\ref{fig:app_qual_text} compares the reproduction of requested text. Ours reduces character errors and improves the completeness of the rendered content in the shown examples, providing qualitative support for the OCR improvements reported in the main text. The final punctuation in example (d) remains imperfect, illustrating a remaining limitation even when most of the target text is correctly reproduced.

\clearpage
\subsubsection{Text Visual Quality}
\label{sec:app_text_visual_quality}

\begin{figure}[H]
    \centering
    \includegraphics[width=\textwidth]{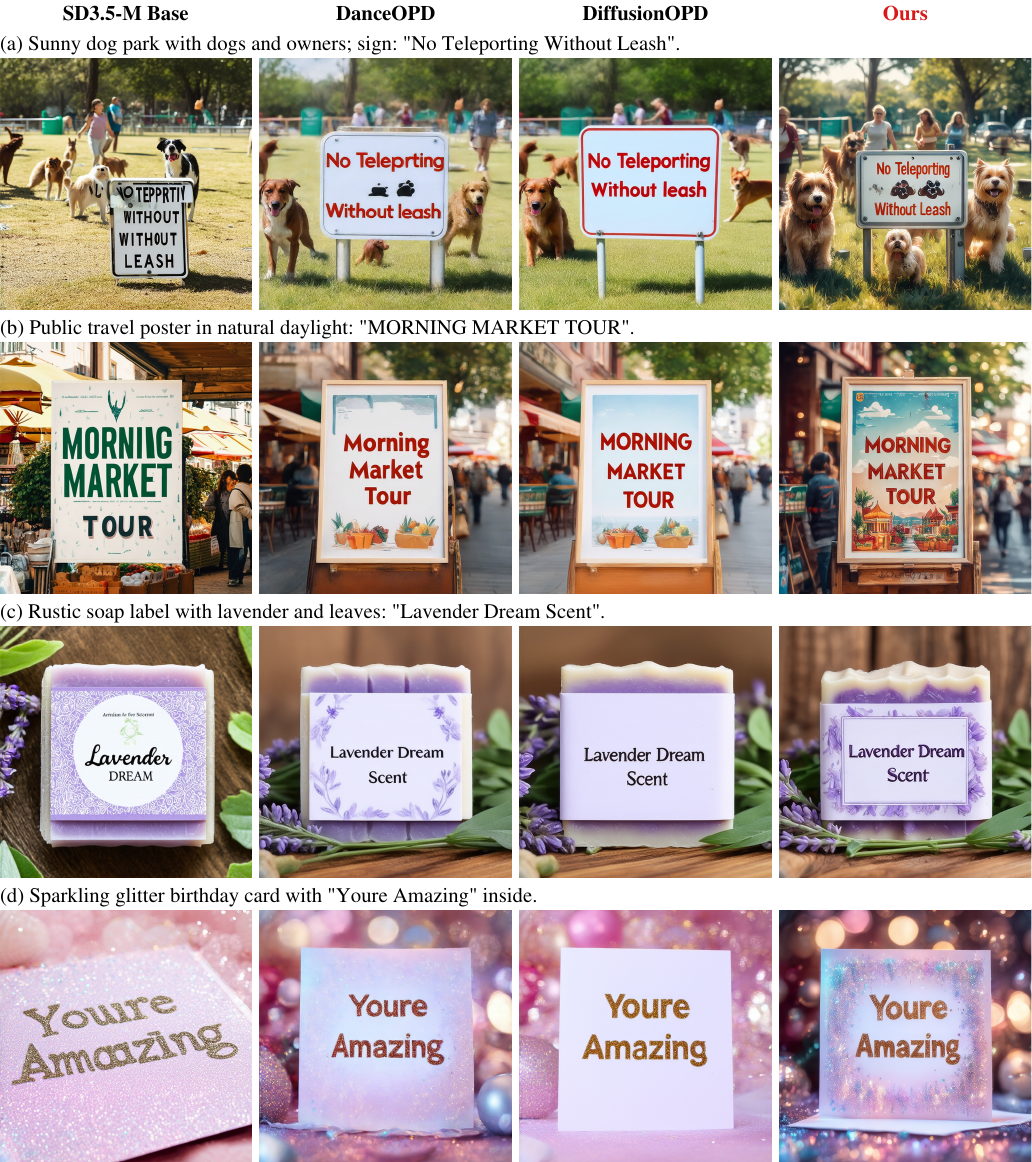}
    \caption{Text visual quality. PhaseOPD unifies readable text, coherent typography, materials, and scene integration.}
    \label{fig:app_qual_text_visual}
\end{figure}

Beyond character accuracy, we examine how generated text fits its visual context. Fig.~\ref{fig:app_qual_text_visual} shows examples in which Ours combines readable lettering with coherent typography, supporting materials, and scene composition. These comparisons complement the OCR scores by illustrating the visual integration of text with the surrounding image.

\clearpage
\subsubsection{Aesthetic Quality}
\label{sec:app_aesthetics}

\begin{figure}[H]
    \centering
    \includegraphics[width=\textwidth]{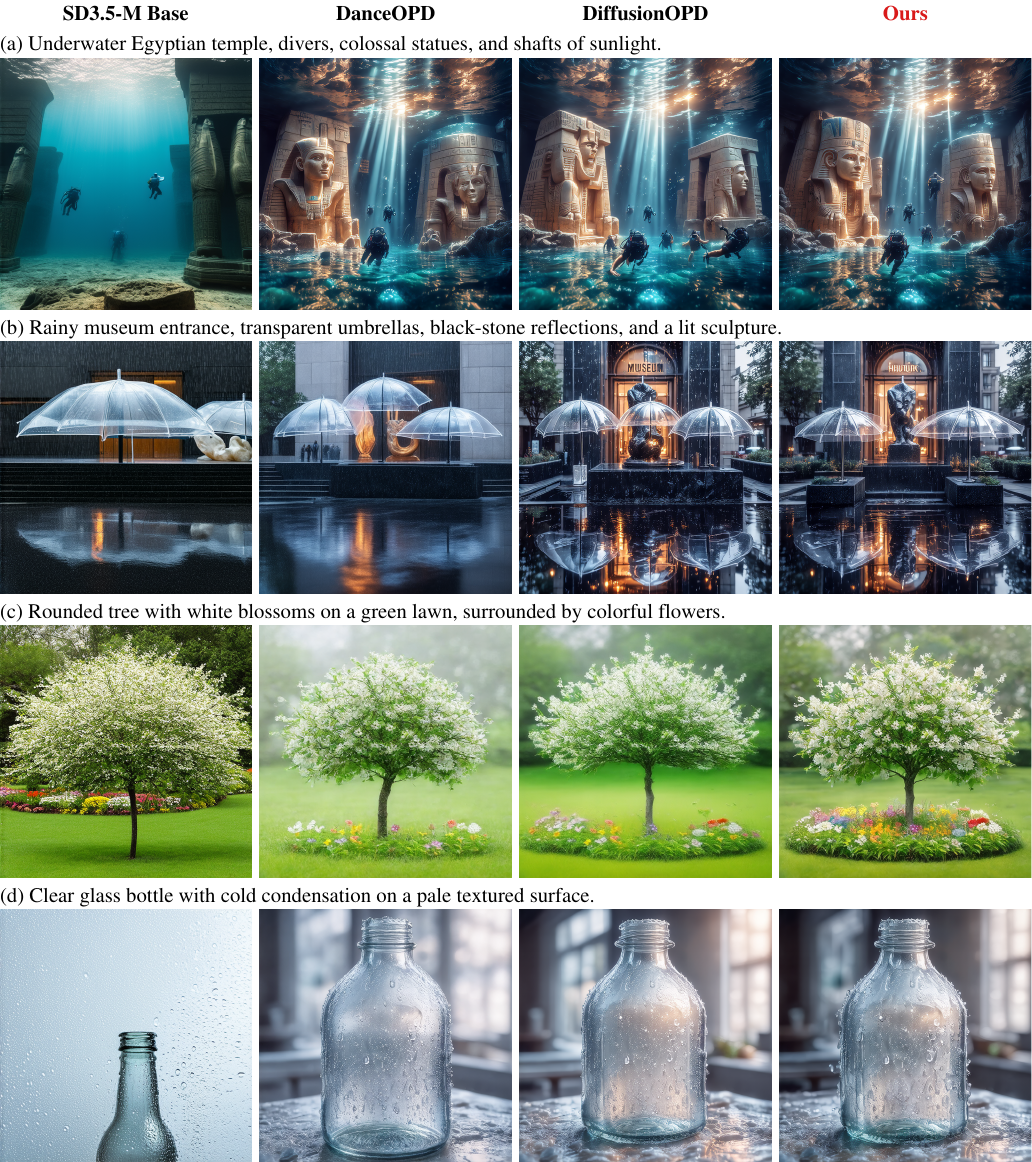}
    \caption{Aesthetic quality. PhaseOPD preserves lighting and materials while clarifying subjects in selected scenes.}
    \label{fig:app_qual_materials}
\end{figure}

Fig.~\ref{fig:app_qual_materials} presents additional comparisons of lighting, materials, and spatial appearance. Ours retains coherent illumination and material detail while producing clearer subject contours and visual separation in selected scenes. Together with the training-cost results, these examples illustrate that the shorter training rollouts can preserve visually detailed outputs.

\clearpage
\subsubsection{Semantic Alignment}
\label{sec:app_semantics}

\begin{figure}[H]
    \centering
    \includegraphics[width=\textwidth]{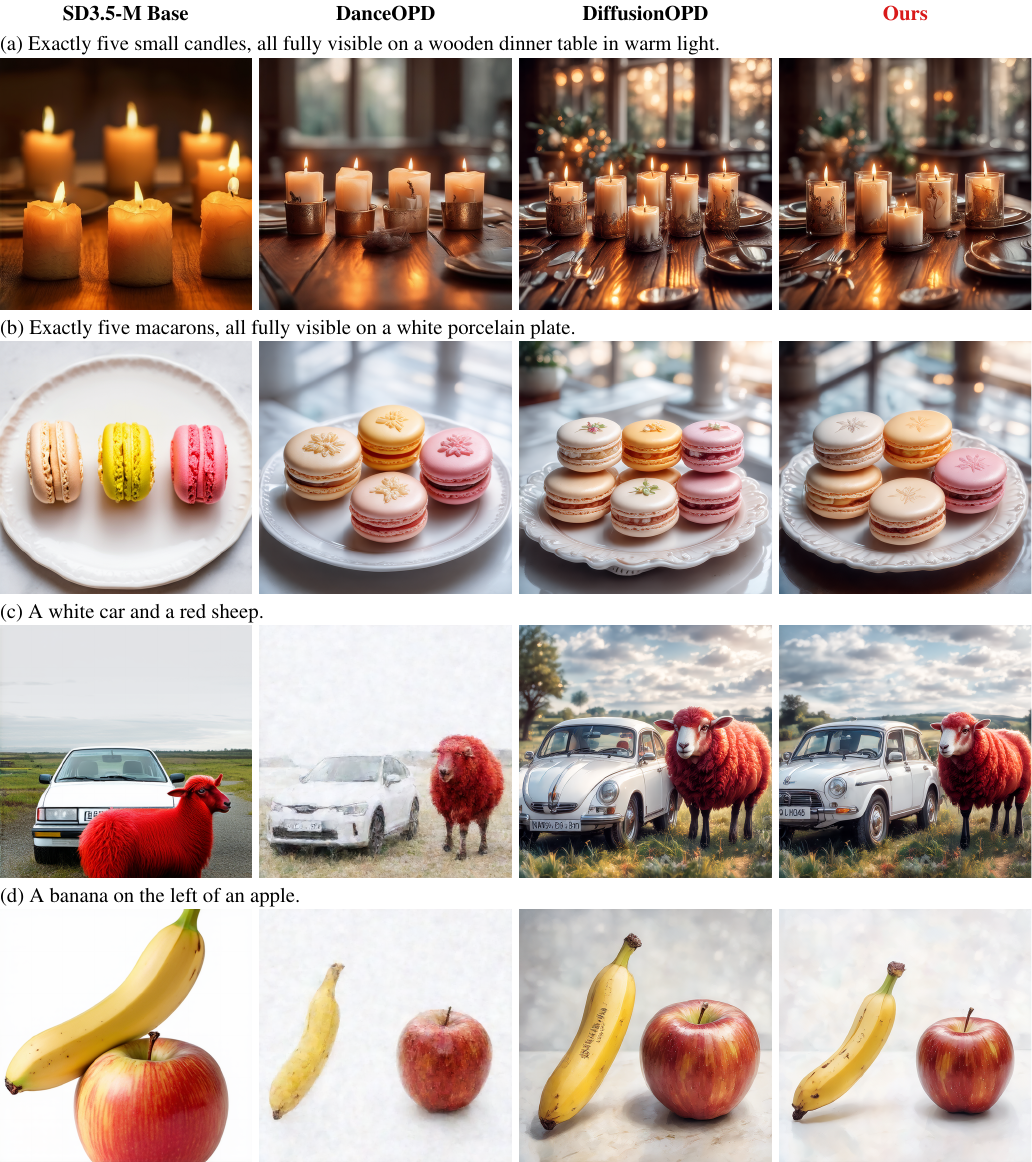}
    \caption{Semantic alignment. PhaseOPD improves counts over DanceOPD in the shown examples while preserving attributes and spatial relations.}
    \label{fig:app_qual_semantic}
\end{figure}

We examine the retention of object counts, attributes, and spatial relations in Fig.~\ref{fig:app_qual_semantic}. Ours improves the shown counting results over DanceOPD while retaining key attributes and scene relationships. These examples complement the GenEval results by showing how compositional improvements appear alongside coherent subject presentation.

\clearpage
\subsubsection{Joint Text and Scene Quality}
\label{sec:app_joint}

\begin{figure}[H]
    \centering
    \includegraphics[width=\textwidth]{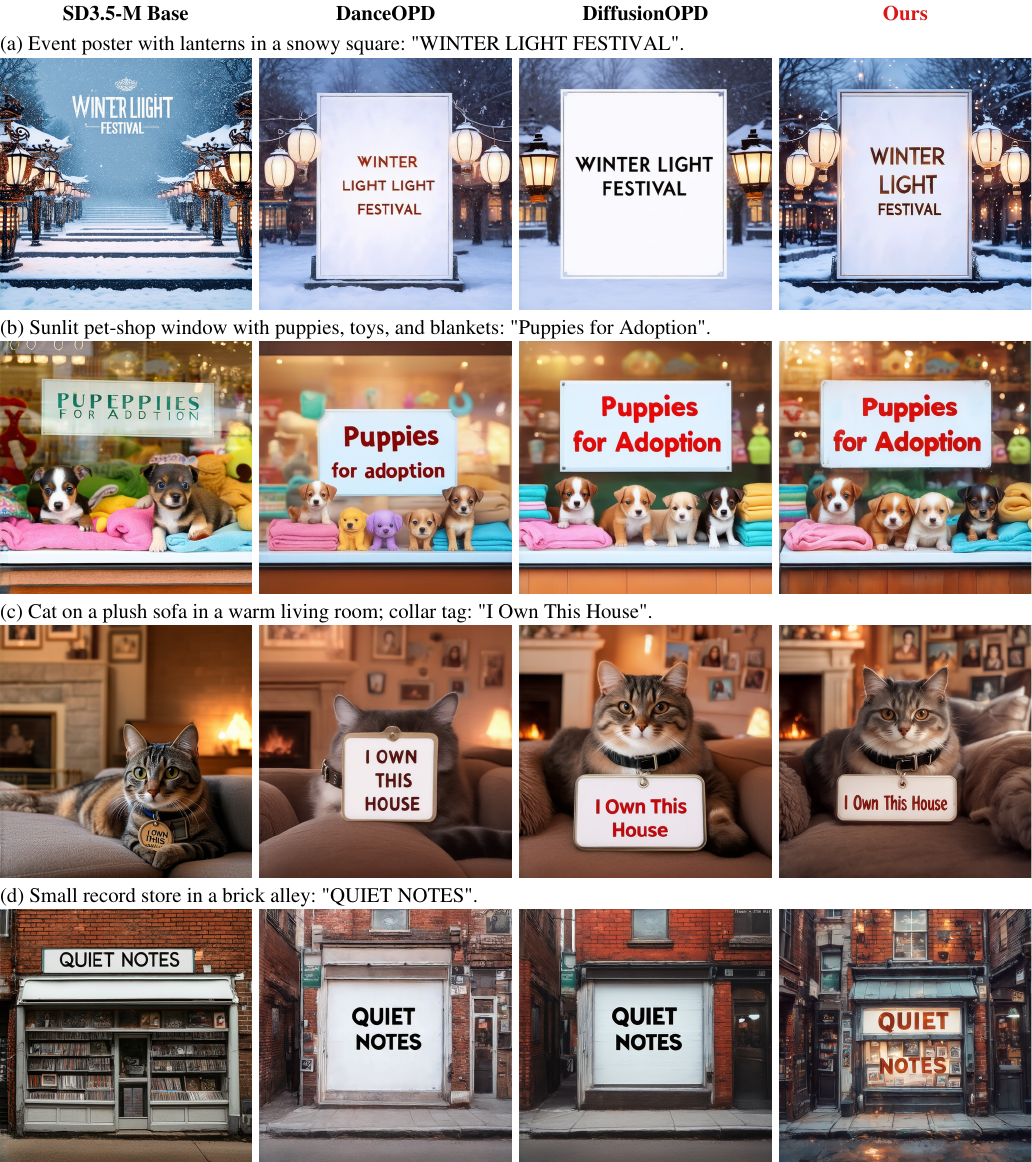}
    \caption{Joint text and scene quality. PhaseOPD combines readable text with clear subjects and coherent surroundings.}
    \label{fig:app_qual_joint}
\end{figure}

Fig.~\ref{fig:app_qual_joint} examines prompts that require readable text and coherent scene content simultaneously. In the shown comparisons, Ours reduces repeated text or subject occlusion and produces more complete text-bearing objects within their surroundings. These examples illustrate how transferred capabilities jointly support local text quality and overall scene organization.

\clearpage

\subsubsection{Alternative Teachers}
\label{sec:app_qual_alternative_teachers}

\begin{figure}[H]
    \centering
    \includegraphics[width=\textwidth]{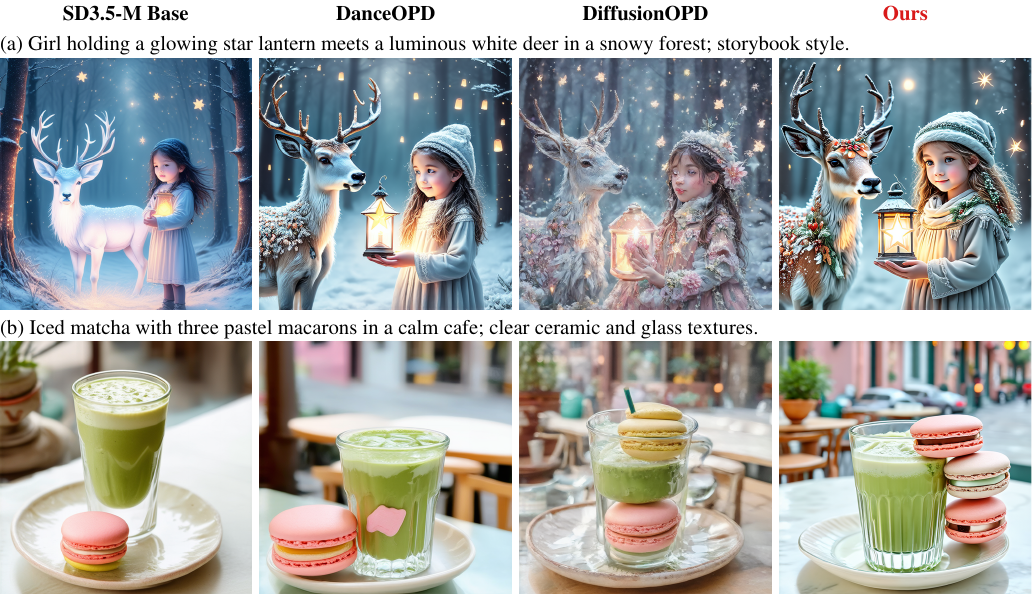}
    \caption{Alternative teachers. PhaseOPD retains scene detail and improves count alignment in the macaron example under HPSv3, SigLIP, and OCR supervision.}
    \label{fig:app_qual_alternate}
\end{figure}

To examine visual transfer beyond the primary teacher set, Fig.~\ref{fig:app_qual_alternate} provides additional results using HPSv3, SigLIP, and OCR teachers. Ours retains subject and scene details and improves count alignment in the macaron example. These observations complement Tab.~\ref{tab:second_teacher_main} by showing that the alternative teacher combination can transfer both appearance and semantic capabilities to the student.

\clearpage
\subsubsection{Cross-Host Transfer}
\label{sec:app_qual_cross_host}

\begin{figure}[H]
    \centering
    \includegraphics[width=\textwidth]{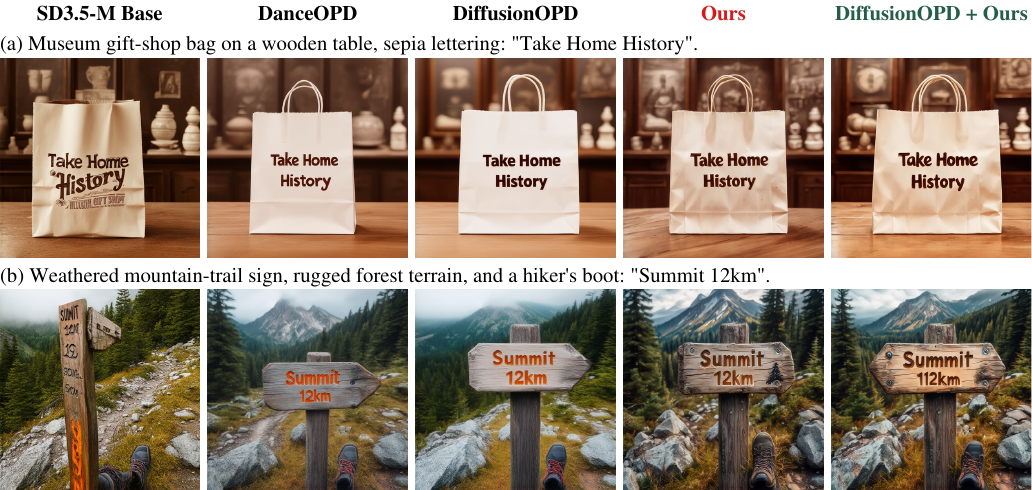}
    \caption{Cross-host transfer. PhaseOPD retains readable text and enriches material appearance with both OPD hosts.}
    \label{fig:appendix_qualitative_cross_host}
\end{figure}

Fig.~\ref{fig:appendix_qualitative_cross_host} compares outputs obtained by adding PhaseOPD to two OPD hosts. In both settings, the augmented models retain readable text and exhibit richer material detail in the shown examples. These comparisons provide qualitative support for the plug-and-play results in Fig.~\ref{fig:plugin_radar_three_host}, illustrating that phase-aware querying can preserve complementary capabilities across different distillation objectives.

\end{CJK*}

\end{CJK*}
\end{document}